\documentclass[runningheads]{llncs}

\usepackage{eccv}

\usepackage{eccvabbrv}

\usepackage{graphicx}
\usepackage{booktabs}

\usepackage[accsupp]{axessibility}  

\usepackage{hyperref}

\usepackage{orcidlink}

\usepackage[table]{xcolor}
\usepackage{multirow}
\usepackage{bbding}
\usepackage{pifont}
\usepackage{amssymb}
\usepackage{wrapfig}

\begin{document}

\title{Incentive Noise and Structural Prior Infusion for Multi-Modal Object Re-Identification} 

\titlerunning{Incentive Noise and Structural Prior for Multi-modal Re-ID}

\author{Weixiang Zhou\inst{1}\textsuperscript{*}\orcidlink{0009-0008-8625-5720} \and
Yuhao Wang\inst{1}\textsuperscript{*}\orcidlink{0009-0000-7906-0329} \and
Xingguo Xu\inst{1}\orcidlink{0009-0008-8286-7367}\and
Weizhen Zhou\inst{2}\orcidlink{0009-0005-3624-0318} \and
Zhixun Su\inst{1}\textsuperscript{\textdagger}\orcidlink{0000-0002-6093-8266} \and
Jinshan Pan\inst{3}\orcidlink{0000-0003-0304-9507} \and
Cong Wang\inst{4}\orcidlink{0000-0002-6068-0103} }


\authorrunning{W.~Zhou et al.}


\institute{
Dalian University of Technology, China \and
New York University, USA \and
Nanjing University of Science and Technology, China \and
University of California, San Francisco, USA \\
\email{\{s20201162006, 924973292, xuxingguo\}@mail.dlut.edu.cn}\\
\email{zxsu@dlut.edu.cn, \{sdluran, supercong94\}@gmail.com}
}

\maketitle

\begingroup
\renewcommand{\thefootnote}{}
\footnotetext{\textsuperscript{*} Equal contribution.}
\footnotetext{\textsuperscript{\textdagger} Corresponding author.}
\endgroup

\begin{abstract}
  Multi-modal object Re-Identification (ReID) benefits from complementary information across heterogeneous imaging modalities. To further enrich semantic representation, text descriptions have recently been incorporated as an additional modality. However, recent vision-language approaches often treat text descriptions as clean, deterministic signals and overlook their inherent noise, including modality-mismatched phrases and semantically ambiguous expressions. Moreover, prevailing methods lack explicit mechanisms to reconcile fine-grained structural discrepancies between modalities, even after high-level semantic alignment. To address these challenges, we propose a novel framework centered on Positive-Incentive Noise ($\pi$-noise) and structured prompt modulation. First, the Semantic Cross-Modal Modulator harnesses task-aware $\pi$-noise, sampled from a distribution conditioned on both visual and text inputs, to perturb global tokens and enable semantics-guided cross-modal compensation. Second, the Structure-Aware Prompt Adapter injects learnable geometric priors via prompts to enhance spatial consistency. Third, the Context-Aware Sparse Fusion module distills structural context to guide adaptive fusion while shielding identity features from noisy local details. Experiments on three multi-modal ReID benchmarks demonstrate the effectiveness and robustness of our approach. The code is available at \url{https://github.com/zw-absin/INSPI}.
\end{abstract}

\begin{figure}[t]
\centering
\includegraphics[width=1.0\linewidth]{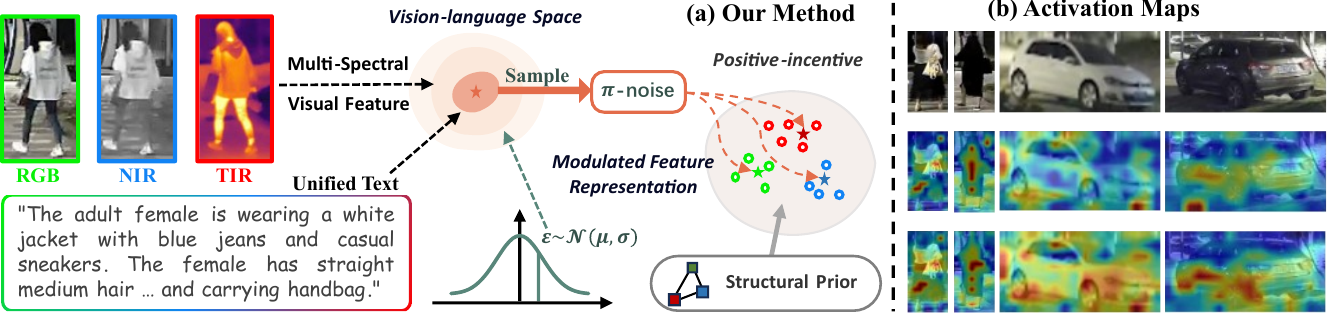}
\caption{
Existing methods rely on uniform local modeling, such as key token selection, but ignore semantic guidance and structural information. 
In contrast, existing text-based methods directly fuse visual and textual features, assuming clean language input, yet propagate noise from ambiguous descriptions. 
    (a) To address these, our method generates task-aware perturbations conditioned on multi-modal inputs, injects learnable structural priors via a shared adapter, and performs context-aware sparse fusion to enhance identity discrimination.
    (b) Feature activation before and after modulation: original images (top), raw activations (middle), enhanced heatmaps (bottom).
}
\label{fig1}
\end{figure}

\section{Introduction}
\label{sec:intro}

Object Re-Identification (ReID) aims to retrieve the same instance across non-overlapping camera views~\cite{zhou2019omni,zhu2020identity,he2021transreid,rao2021counterfactual,somers2024keypoint}.  
Most existing single-modality approaches rely solely on RGB imagery.  
These methods often perform poorly under adverse imaging conditions, such as occlusion, low illumination, or motion blur, where discriminative visual cues are significantly degraded or lost.
To mitigate this limitation, multi-modal object ReID~\cite{wang2022interact,zheng2023dynamic,wu2025lrmm,li2025video,li2025shape,feng2025multi,li2026causal} has emerged as a promising direction by leveraging complementary information from multiple imaging modalities, such as infrared, depth, or event-based sensors, to learn more robust and discriminative feature representations in complex scenarios.

Recent advances in Contrastive Language-Image Pre-training (CLIP)~\cite{radford2021learning} have further propelled multi-modal ReID by incorporating textual semantics~\cite{li2025icpl,lin2025dmpt}.
Moreover, the rise of foundation models and large language models has inspired new paradigms that incorporate auxiliary semantic signals, such as object masks or text descriptions, into ReID pipelines~\cite{wang2025idea,li2025next,xu2026stmi,gao2026chain} to enrich identity representation and facilitate semantic-aware cross-modal alignment.  
These methods expand the modality space and enhance robustness under degraded visual conditions. For instance, Wang et al.~\cite{wang2025idea} integrate inverted text sequences into visual features via deformable aggregation, while Xu et al.~\cite{xu2026stmi} jointly leverage mask and text guidance for cross-modal modeling.  
Despite their effectiveness, these approaches treat text inputs as clean and deterministic anchors and ignore the inherent noise in natural language descriptions. Such noise includes modality-mismatched phrases, irrelevant padding tokens, and semantically ambiguous expressions, all of which can introduce harmful biases into representation learning and undermine alignment reliability.

Separately, multi-modal ReID faces another fundamental challenge arising from the heterogeneity of imaging sensors. The diversity across spectral modalities enables complementary perception but also introduces substantial background clutter from modality-specific artifacts and pixel-level inter-modal misalignment due to heterogeneous sensor characteristics~\cite{liu2025signal}.
Existing methods address these issues primarily through two strategies. The first refines feature representations via selective regional sampling, emphasizing discriminative regions and suppressing unreliable ones~\cite{zhang2024magic,wan2025reliable}. The second models modality-specific properties to preserve intrinsic discriminability while encouraging cross-modal complementarity~\cite{wang2024top,wang2025decoupled}.
For example, Zhang et al.~\cite{zhang2024magic} suppress background interference through token-level attention masking, while Wang et al.~\cite{wang2024top} decouple modality-specific features to retain individual characteristics.  
Nevertheless, these approaches typically assume uniform alignment across modalities and rely heavily on global constraints or model adaptivity, thereby overlooking fine-grained structural discrepancies that persist even after high-level semantic alignment.

To address these limitations, we propose a novel framework for multi-modal object ReID, centered on Positive-Incentive Noise ($\pi$-noise)~\cite{li2022positive,zhang2025variational} and structured prompt modulation, as illustrated in Figure~\ref{fig1}.
Our core insight is that noisy textual semantics can serve as a source of beneficial uncertainty when properly regularized.  
In contrast to prior work~\cite{feng2025multi} that suppresses unreliable visual regions by modeling visual uncertainty, our approach harnesses textual uncertainty to enhance feature learning.
Specifically, we generate a task-aware noise distribution conditioned on both visual content and text captions. A sampled perturbation from this distribution is injected into the global token, yielding a semantically reinforced representation. This $\pi$-noise mechanism not only mitigates the adverse effects of textual noise but also actively promotes identity-discriminative feature refinement.
Building on this principle, our framework integrates three synergistic components.  
First, the Semantic Cross-Modal Modulator (SCM) leverages $\pi$-noise-enhanced tokens as dynamic controllers to enable semantics-guided alignment and cross-modal complementarity.  
Second, the Structure-Aware Prompt Adapter (SPA) initializes modality-specific structural prompts to preserve intrinsic geometric cues and aligns them through a shared adapter to encourage cross-modal structural consistency.  
Third, the Context-Aware Sparse Fusion (CASF) distills self-supervised structural knowledge into a clean contextual signal, which governs sparse expert routing. This design achieves adaptive fusion while shielding identity features from direct exposure to noisy local details.
Extensive experiments on three multi-modal ReID benchmarks demonstrate the effectiveness and generalization capability of our approach.
Our main contributions are summarized as follows:
\begin{itemize}
    \item We introduce Positive-Incentive Noise ($\pi$-noise), a novel representation learning paradigm that transforms noisy textual semantics into a beneficial perturbation for semantic reinforcement in multi-modal object ReID.
    \item We propose a unified multi-modal ReID framework integrating three components: the Semantic Cross-Modal Modulator (SCM) leveraging $\pi$-noise for semantic modulation; the Structure-Aware Prompt Adapter (SPA) injecting geometric priors via learnable prompts; and the Context-Aware Sparse Fusion (CASF) performing noise-resilient fusion via expert routing.
    \item Extensive experiments on three public multi-modal object ReID datasets, including RGBNT201, RGBNT100, and MSVR310, demonstrate the superiority of our method over state-of-the-art approaches.
\end{itemize}

\section{Related Work}

\subsection{Multi-Modal Object Re-Identification}
Multi-modal object ReID aims to match objects across camera views by fusing complementary cues from heterogeneous modalities. Unlike single-modality approaches, recent methods emphasize cross-modal interaction mechanisms that preserve modality-specific characteristics while enforcing semantic consistency~\cite{zhang2025prompt,wang2025mambapro,feng2025mdreid,wang2025decoupled,yang2025dinov2}. For example, Wang et al.~\cite{wang2025mambapro} use Mamba to model intra- and inter-modal dependencies. Feng et al.~\cite{feng2025multi} embed a Mixture-of-Experts (MoE) module in Vision Transformers to capture modality-specific features. Wang et al.~\cite{wang2025decoupled} further decouple features under varying visual quality. 
Concurrently, alternative interaction strategies such as graph-based reasoning~\cite{wan2026mgrnet}, parameter-efficient tuning~\cite{li2026peftboa}, and hierarchical token alignment~\cite{liu2026signal} have also been explored.
Beyond fusing visual modalities, recent work has explored leveraging textual semantics to further bridge the semantic gap and provide high-level identity priors.
The emergence of vision-language models like CLIP~\cite{radford2021learning} has motivated the integration of text descriptions into multi-modal object ReID to enrich semantic context and align cross-modal representations~\cite{lin2025dmpt,wang2025idea,li2025next,xu2026stmi}.
These approaches typically treat text as a clean semantic anchor. However, they often ignore the inherent noise in text annotations, such as ambiguity, incompleteness, or misalignment, which can degrade representation learning when fused directly.
To address this issue, our method embraces textual noise as a source of beneficial uncertainty through Positive-Incentive Noise, which actively promotes semantic reinforcement.

\subsection{The Constructive Role of Noise}
Contrary to the conventional assumption that noise is always harmful, recent studies suggest that certain types of noise can be beneficial. The concept of Positive-Incentive Noise was introduced to characterize perturbations that reduce task entropy, thereby simplifying the learning problem~\cite{li2022positive,zhang2025variational}. Under this framework, even random noise may act as $\pi$-noise if it lowers the effective complexity of a task. This perspective has inspired methods that leverage noise as a regularizer or structural guide. For instance, Huang et al.~\cite{huang2025enhance} use structured noise to improve vision-language alignment. Jiang et al.~\cite{jiang2025mixture} propose a mixture-of-noise mechanism for class incremental learning, where task-specific noise is injected into intermediate features to suppress irrelevant patterns and mitigate parameter drift in pre-trained backbones. Together, these works demonstrate that properly designed noise can serve as a constructive inductive bias. 
Nevertheless, the potential of $\pi$-noise in multi-modal object ReID, especially for harnessing noisy text inputs, has not been explored.
In this work, we bridge this gap by generating $\pi$-noise through cross-modal fusion of visual and textual semantics and leveraging it to enhance feature representations.

\begin{figure}[t]
\centering
\includegraphics[width=\linewidth]{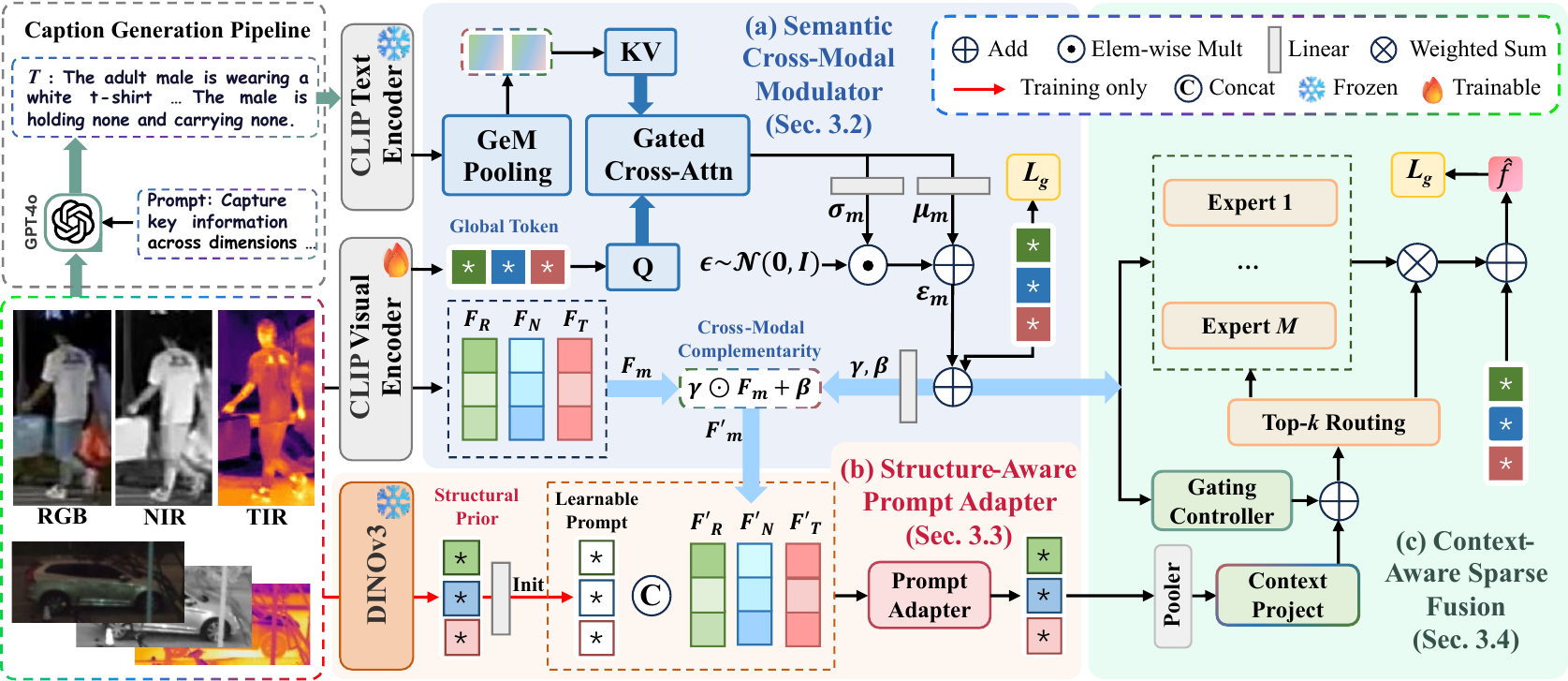}
\caption{Overview of the proposed framework.
(a) The Semantic Cross-Modal Modulator (SCM) samples beneficial perturbations from a distribution conditioned on both visual and text inputs to enable semantics-guided cross-modal compensation. (b) The Structure-Aware Prompt Adapter (SPA) injects learnable structural priors to promote cross-modal structural alignment. (c) The Context-Aware Sparse Fusion (CASF) module performs multi-modal feature fusion via sparse routing guided by structural context.
With these components, our method achieves robust identity matching under noisy text descriptions and heterogeneous imaging conditions.
}
\label{fig:framework}
\end{figure}

\section{Methodology}\label{sec:method}
As illustrated in Figure~\ref{fig:framework}, we propose a novel framework for multi-modal object ReID comprising three core components. 
First, the Semantic Cross-Modal Modulator (SCM) introduces the concept of Positive-Incentive Noise ($\pi$-noise) into ReID by sampling task-aware perturbations from a distribution conditioned on both visual and text inputs, thereby enhancing global representations and enabling semantics-guided cross-modal compensation. 
Second, the Structure-Aware Prompt Adapter (SPA) injects modality-specific, learnable structural priors to enrich geometric awareness while promoting cross-modal structural alignment through a shared adapter. 
Third, the Context-Aware Sparse Fusion (CASF) module leverages structural context to modulate sparse routing during multi-modal feature fusion. 
We detail each component of our framework below.

\subsection{Feature Initialization}
Following recent vision-language ReID approaches~\cite{wang2025idea,xu2026stmi}, we employ the pre-trained CLIP encoder~\cite{radford2021learning} to extract features from multi-spectral images and modality-agnostic text captions. 
This yields visual features $F_v = \{f_m, F_m\}$ and text features $F_t$, where $m \in \{R, N, T\}$ denotes the RGB, Near-Infrared (NIR), and Thermal Infrared (TIR) modalities, respectively. 
For each modality $m$, $f_m \in \mathbb{R}^{D}$ represents the visual global token, and $F_m \in \mathbb{R}^{N_p \times D}$ denotes the sequence of $N_p$ visual patch tokens. 
The text input is encoded as $F_t \in \mathbb{R}^{N_t \times D}$, where $N_t$ is the number of text tokens.
Then, these features are fed into our proposed modules for further refinement.

\subsection{Semantic Cross-Modal Modulator}
To mitigate the adverse effects of noisy text cues while enabling semantic-aware cross-modal alignment and complementarity, we propose the Semantic Cross-Modal Modulator (SCM). Inspired by the concept of  Positive-Incentive Noise ($\pi$-noise)~\cite{li2022positive,huang2025enhance}, we reinterpret noise in representation learning. Specifically, by leveraging both textual and visual features and applying proper regularization techniques, we can generate a structured perturbation. This perturbation is designed to reinforce identity-discriminative semantics, turning potential noise into a positive incentive for enhancing cross-modal alignment and complementarity.

Specifically, as illustrated in Figure~\ref{fig:framework}(a), the input text sequence is first condensed into a global text feature $f_t \in \mathbb{R}^{N'_t \times D}$ via GeM pooling~\cite{radenovic2018fine} after removing padding tokens. For each available visual modality $m \in \{R, N, T\}$ with global token $f_m \in \mathbb{R}^D$, we employ a Gated Attention-based Cross-Attention (GCA) module~\cite{qiu2025gated} to fuse textual and visual semantics with the following equation:
\begin{equation}
    f'_m = \mathrm{GCA}(f_m, f_t) \in \mathbb{R}^D.
\end{equation}
This fused representation is then projected to obtain a Gaussian distribution:
\begin{equation}
    \mu_m = L_{\mu}(f'_m), \quad \sigma_m = L_{\sigma}(f'_m) \in \mathbb{R}^D,
\end{equation}
where $\mu_m$ represents the mean vector, and $\sigma_m$ denotes the standard deviation vector.
Then, we sample structured noise via the reparameterization trick:
\begin{equation}
    \varepsilon_m = \mu_m + \sigma_m \odot \epsilon, \quad \epsilon \sim \mathcal{N}(0, I),
\end{equation}
where $\epsilon$ is a standard normal random vector. Crucially, this perturbation is not random but semantically conditioned. 
It is shaped by both visual content and textual semantics, actively reshaping features toward identity-discriminative structures. 
As a result, it establishes an information-gain mechanism rather than introducing task-irrelevant variations.

The resulting semantically enhanced global token is constructed as:
\begin{equation}
    \tilde{f}_m = f_m + \varepsilon_m \in \mathbb{R}^D.
\end{equation}
These enhanced tokens $\tilde{f}_m$ serve as semantic controllers for cross-modal modulation. 
Given a source modality's local features $F_i \in \mathbb{R}^{N_p \times D}$ and a target modality's enhanced token $\tilde{f}_j$, SCM applies an affine transformation to $F_i$:
\begin{equation}
    \gamma_j = \mathrm{MLP}_{\gamma}(\tilde{f}_j), \quad \beta_j = \mathrm{MLP}_{\beta}(\tilde{f}_j) \in \mathbb{R}^D,
\end{equation}
\begin{equation}
    F'_i = F_i \odot \gamma_j + \beta_j \in \mathbb{R}^{N_p \times D}, \quad i,j \in \{R, N, T\}.
\end{equation}
This operation reinterprets the content of modality $i$ through the semantic lens of modality $j$, enabling semantics-guided feature compensation.
Notably, since modulation relies only on the enhanced tokens of available modalities, our design naturally supports robust inference under arbitrary modality-missing patterns.

\subsection{Structure-Aware Prompt Adapter}
In multi-modal ReID, multi-spectral images exhibit substantial appearance discrepancies yet share consistent high-level structural semantics. However, prevailing vision encoders such as CLIP, while offering strong discriminative power, inadequately model cross-modal structural consistency, particularly in non-visible modalities where geometric priors are prone to degradation.
This limitation arises from the training objective of large-scale contrastive models, which emphasizes global semantic alignment but often sacrifices fine-grained structural fidelity in dense representations~\cite{simeoni2025dinov3}.

To address this issue, we propose the Structure-Aware Prompt Adapter (SPA), a training-inference decoupled module that injects structure-aware global priors into learnable prompts to guide cross-modal feature alignment and interaction within a unified semantic space.
Specifically, during training, we initialize modality-specific learnable prompts using the global tokens extracted from DINOv3~\cite{simeoni2025dinov3}, which are known to encode rich structural information. 
Critically, DINOv3 is used only once at initialization and is entirely detached during both training and inference. This design leverages the frozen model’s superior structural awareness without incurring additional computational overhead. As optimization proceeds, these prompts internalize structural priors and serve dual roles: acting as modality identifiers and structural anchors.
These prompts are then linearly projected into the CLIP feature space and prepended to the visual token sequence $F'_m$ of each modality, forming an augmented sequence:
\begin{equation}
    \overline{F}_m = [\tilde{f}_m, F'_m] \in \mathbb{R}^{(N_p+1) \times D},
\end{equation}
where $[\cdot]$ denotes concatenation. The three modality-specific sequences are then stacked into a unified input and processed by a shared adapter block $\mathcal{A}$ as follows:
\begin{equation}
    \overline{F} = [\overline{F}_R, \overline{F}_N, \overline{F}_T],
    \quad
    \hat{F} = \mathcal{A}(\overline{F}) \in \mathbb{R}^{3(N_p+1) \times D},
\end{equation}
where $\mathcal{A}$ consists of Multi-Head Self-Attention (MHSA) and a Feed-Forward Network (FFN)~\cite{dosovitskiy2020image}.
This shared adapter jointly performs cross-modal fusion and dynamic prompt adaptation, producing context-rich representations that preserve structural consistency while enabling semantic interaction across modalities.

\subsection{Context-Aware Sparse Fusion}
Omitting structural modeling leads to the loss of geometric and contour cues, while directly fusing raw local features risks contamination from low-quality images and compromises robustness.
To reconcile this dilemma, we propose Context-Aware Sparse Fusion (CASF), which distills self-supervised structural priors into a compact context representation that is both structure-aware and semantically coherent. This context does not contribute directly to the final identity embedding. Instead, it serves as a routing signal to dynamically modulate the fusion of multi-modal global tokens.

Specifically, we first extract the global components $\hat{F}_g$ from $\hat{F}$ and apply average pooling $\mathcal{P}$ to obtain a coarse structural summary.
The context signal $\hat{f}_c \in \mathbb{R}^D$ is then computed as:
\begin{equation}
    \hat{f}_c = L_{\text{CP}}(\mathcal{P}(\hat{F}_g)) + L_{\text{GC}}(\tilde{f}_g),
    \quad
    \tilde{f}_g = [\tilde{f}_R, \tilde{f}_N, \tilde{f}_T],
\end{equation}
where $L_{\text{CP}}$ (Context Projection) and $L_{\text{GC}}$ (Gating Controller) are linear projections.
This context modulates the gating logits of a sparsely-gated Mixture-of-Experts (MoE) layer~\cite{shazeer2017outrageously}. The input to the MoE is the global tokens $\tilde{f}_g \in \mathbb{R}^{3D}$. Only the top-$k$ experts are activated per sample, and their outputs are combined via learned weights to produce a fused representation $f_{\text{fuse}} \in \mathbb{R}^D$. 
Because expert selection is guided by structural context rather than raw features, the fusion process becomes inherently resilient to noise and modality degradation. Moreover, since the identity representation is constructed without direct exposure to local details, CASF achieves a principled balance between discriminability and robustness.
The final discriminative feature for each modality $m \in \{R, N, T\}$ is obtained by residual addition:
\begin{equation}
    \hat{f}_m = f_m + f_{\text{fuse}},
    \quad
    \hat{f} = [\hat{f}_R,\hat{f}_N,\hat{f}_T].
\end{equation}
%

%
\subsection{Objective Functions}

As shown in Figure~\ref{fig:framework}, the network is trained with the following loss function:
\begin{equation}
    \mathcal{L} = \mathcal{L}_g([f_{R}, f_{N}, f_{T}]) + \mathcal{L}_g(\hat{f}) + 0.01 \cdot \mathcal{L}_{\mathrm{aux}},
\end{equation}
where $\mathcal{L}_g(\cdot)$ denotes the sum of label smoothing cross-entropy loss~\cite{szegedy2016rethinking} and triplet loss~\cite{hermans2017defense}. 
This formulation applies $\mathcal{L}_g$ separately to the modality-specific features $[f_R, f_N, f_T]$ and to the CASF-fused representation $\hat{f}$, providing dual supervision that promotes discriminative representations at both stages. 
$\mathcal{L}_{\mathrm{aux}}$ is an auxiliary load-balancing loss from the MoE layer, which encourages balanced expert utilization. The above losses ensure effective joint optimization of feature learning. 
\section{Experiments}
\subsection{Datasets and Evaluation Metrics}
We evaluate our method on 3 public multi-modal ReID benchmarks. 
RGBNT201 \cite{zheng2021robust} is a large-scale person ReID dataset containing 201 identities captured across four non-overlapping cameras with synchronized RGB, Near-Infrared (NIR), and Thermal Infrared (TIR) images for each identity. It includes 4787 aligned images per modality and exhibits significant challenges such as viewpoint changes, occlusion, cluttered backgrounds, and imaging degradations due to adverse lighting or weather conditions. For vehicle ReID, we use RGBNT100~\cite{li2020multi}, which comprises 17250 aligned RGB–NIR–TIR triplets across 100 vehicles and features viewpoint variation, low illumination, and partial occlusion. Additionally, we include MSVR310~\cite{zheng2023cross}, a compact but highly challenging vehicle dataset with 2087 triplets collected under extreme conditions, making it a stringent test of robustness.
Following standard practice in cross-modal person ReID, we report mean average precision (mAP) and Rank-$K$ accuracy for $K \in \{1, 5, 10\}$.

\begin{table}[t]
\centering
\setlength\tabcolsep{1.5pt}
\fontsize{7}{7}\selectfont 
\renewcommand{\arraystretch}{1.4}
\begin{minipage}[t]{0.47\linewidth}%
\centering
\setlength\tabcolsep{0.8pt}%
\caption{Performance comparison on RGBNT201. The best and second-best results are shown in \textbf{bold} and \underline{underlined}, respectively. Methods with $\dagger$ are CLIP-based, those with $*$ are ViT-based, and the rest are CNN-based.}
\label{tab:multi-spectral person ReID}
\begin{tabular}[t]{cr|cccc}
\noalign{\hrule height 1pt}
&\multicolumn{1}{c|}{\multirow{2}{*}{\textbf{Methods}}} & \multicolumn{4}{c}{\textbf{RGBNT201}} \\ \cline{3-6}
& & \textbf{mAP} & \textbf{R-1} & \textbf{R-5} & \textbf{R-10} \\ \hline
\multirow{3}{*}{\rotatebox{90}{\textbf{Single}}}
 &OSNet$_{2019}$~\cite{zhou2019omni}  & 25.4 & 22.3 & 35.1 & 44.7 \\
 &CAL$_{2021}$~\cite{rao2021counterfactual}  & 27.6 & 24.3 & 36.5 & 45.7 \\
 &PCB$_{2018}$~\cite{sun2018beyond}  & 32.8 & 28.1 & 37.4 & 46.9 \\ \hline
\multirow{16}{*}{\rotatebox{90}{\textbf{Multi-Modal}}}
 & HAMNet$_{2020}$~\cite{li2020multi}   & 27.7 & 26.3 & 41.5 & 51.7 \\
 & PFNet$_{2021}$~\cite{zheng2021robust}    & 38.5 & 38.9 & 52.0 & 58.4 \\
 & IEEE$_{2022}$~\cite{wang2022interact}     & 47.5 & 44.4 & 57.1 & 63.6 \\
 & DENet$_{2023}$~\cite{zheng2023dynamic}    & 42.4 & 42.2 & 55.3 & 64.5 \\
 & LRMM$_{2025}$~\cite{wu2025lrmm} & 52.3 & 53.4 & 64.6 & 73.2\\
 & UniCat$^*_{2023}$~\cite{crawford2023unicat}   & 57.0 & 55.7 & - & - \\
& HTT$^*_{2024}$~\cite{wang2024heterogeneous} &71.1 &73.4 &83.1 &87.3\\
& TOP-ReID$^*_{2024}$~\cite{wang2024top}  &72.3 &76.6 &84.7 &89.4\\
& EDITOR$^*_{2024}$~\cite{zhang2024magic} & 66.5 & 68.3 & 81.1 & 88.2 \\
& RSCNet$^*_{2024}$~\cite{yu2024representation} & 68.2 & 72.5 & - & - \\
& WTSF-ReID$^*_{2025}$~\cite{yu2025wtsf} & 67.9 &72.2 &83.4 &89.7 \\
& DESANet$^*_{2025}$~\cite{dong2025escaping} & 74.6 &77.6 &87.1 &91.3 \\
& PromptMA$^\dagger_{2025}$~\cite{zhang2025prompt} & 78.4 &80.9 &87.0 &88.9 \\
& MambaPro$^\dagger_{2025}$~\cite{wang2025mambapro} & 78.9 & \underline{83.4} & {89.8} & 91.9 \\
& DeMo$^\dagger_{2025}$~\cite{wang2025decoupled}  &{79.0} &{82.3} &88.8 &{92.0} \\
& IDEA$^\dagger_{2025}$~\cite{wang2025idea}  &\underline{80.2} &82.1 &\underline{90.0} &\underline{93.3} \\
\rowcolor[gray]{0.92}
 & ${\textbf{Ours}}^\dagger$  &\textbf{80.6} &\textbf{83.9} &\textbf{91.6} &\textbf{93.4} \\
\noalign{\hrule height 1pt}
\end{tabular}
\end{minipage}
\hfill
\begin{minipage}[t]{0.51\linewidth}%
\centering
\setlength\tabcolsep{0.9pt}%
\caption{Performance comparison on RGBNT100 and MSVR310.}
\vspace{+1.7mm}
\label{tab:multi-spectral vehicle ReID}
\begin{tabular}[t]{cr|cccc}
\noalign{\hrule height 1pt}
&\multicolumn{1}{c|}{\multirow{2}{*}{\textbf{Methods}}} & \multicolumn{2}{c}{\textbf{RGBNT100}} & \multicolumn{2}{c}{\textbf{MSVR310}} \\ \cline{3-6}
& & \textbf{mAP} & \textbf{R-1} & \textbf{mAP} & \textbf{R-1} \\ \hline
\multirow{3}{*}{\rotatebox{90}{\textbf{Single}}}
&PCB$_{2018}$~\cite{sun2018beyond}& 57.2 & 83.5 & 23.2 & 42.9 \\
&OSNet$_{2019}$~\cite{zhou2019omni}& 75.0 & 95.6 & 28.7 & 44.8 \\
&TransReID$^*_{2021}$~\cite{he2021transreid}& 75.6 & 92.9 & 18.4 & 29.6 \\ \hline
\multirow{18}{*}{\rotatebox{90}{\textbf{Multi-Modal}}}
&HAMNet$_{2020}$~\cite{li2020multi} & 74.5 & 93.3 & 27.1 & 42.3 \\
&PFNet$_{2021}$~\cite{zheng2021robust}& 68.1 & 94.1 & 23.5 & 37.4 \\
&GAFNet$_{2022}$~\cite{guo2022generative} & 74.4 & 93.4 & - & - \\
&GPFNet$_{2023}$~\cite{he2023graph} & 75.0 & 94.5 & - & - \\
&CCNet$_{2023}$~\cite{zheng2023cross} & 77.2 & 96.3 & 36.4 & 55.2 \\
& LRMM$_{2025}$~\cite{wu2025lrmm} & 78.6 & 96.7 & 36.7 &49.7\\
&GraFT$^*_{2023}$~\cite{yin2023graft} &76.6 &94.3 &- &-\\
&UniCat$^*_{2023}$~\cite{crawford2023unicat}    & 79.4 & 96.2 & - & - \\
&PHT$^*_{2023}$~\cite{pan2023progressively} & 79.9 & 92.7 & - & - \\
& HTT$^*_{2024}$~\cite{wang2024heterogeneous} &75.7&92.6&- &-\\
& TOP-ReID$^*_{2024}$~\cite{wang2024top} &81.2 & 96.4 & 35.9 & 44.6 \\
& EDITOR$^*_{2024}$~\cite{zhang2024magic} & 82.1 & 96.4 &39.0 & 49.3\\
& RSCNet$^*_{2024}$~\cite{yu2024representation} &82.3 &96.6 &39.5 &49.6\\
& FACENet$^*_{2025}$~\cite{zheng2025flare} & 81.5 &{96.9} &36.2 &54.1 \\
& WTSF-ReID$^*_{2025}$~\cite{yu2025wtsf} & 82.2 &96.5 & 39.2 & 49.1 \\
& DESANet$^*_{2025}$~\cite{dong2025escaping} & 82.1 &97.4 &39.2 &47.8 \\
& PromptMA$^\dagger_{2025}$~\cite{zhang2025prompt} & 85.3 & {97.4} &{55.2} &{64.5} \\
& MambaPro$^\dagger_{2025}$~\cite{wang2025mambapro} & 83.9 & 94.7 &{47.0} & 56.5 \\
& DeMo$^\dagger_{2025}$~\cite{wang2025decoupled} &{86.2} &\underline{97.6} &\underline{49.2} &{59.8} \\
& IDEA$^\dagger_{2025}$~\cite{wang2025idea}& \underline{87.2} &96.5 &{47.0} &\underline{62.4} \\
\rowcolor[gray]{0.92}
& ${\textbf{Ours}}^\dagger$&\textbf{89.9} &\textbf{98.2} &\textbf{65.0} &\textbf{77.2} \\
\noalign{\hrule height 1pt}
\end{tabular}
\end{minipage}
\end{table}

\subsection{Implementation Details}
Our framework is implemented in PyTorch and trained on an NVIDIA A6000 GPU, using CLIP~\cite{radford2021learning} as both the visual and text backbone. Input images are resized to $256 \times 128$ for RGBNT201 and to $128 \times 256$ for RGBNT100 and MSVR310. Data augmentation includes random horizontal flipping, random cropping, and random erasing~\cite{zhong2020random} to enhance model robustness. The mini-batch size is set to 64 for RGBNT201 and MSVR310, and 128 for RGBNT100, with 8 images per identity for RGBNT201 and MSVR310 and 16 images per identity for RGBNT100. All models are trained for 50 epochs. We use the Adam optimizer to fine-tune all learnable parameters, starting with an initial learning rate of $3.5 \times 10^{-6}$, which is linearly decayed to $3.5 \times 10^{-7}$ over the course of training. Text prompts are generated using the caption generation pipeline proposed by Xu et al.~\cite{xu2026stmi}. 
\subsection{Comparison with State-of-the-Art Methods}
\textbf{Multi-Modal Person ReID.}
Table~\ref{tab:multi-spectral person ReID} compares our method (Ours$^\dagger$) with existing multi-modal approaches on RGBNT201.
The results confirm that leveraging complementary modalities consistently improves upon both single-modality baselines and CNN-based methods. Our method achieves 80.6\% mAP and 83.9\% Rank-1 accuracy, outperforming TOP-ReID$^*$ by 8.3\% in mAP and 7.3\% in Rank-1. Notably, compared to recent CLIP-based methods such as MambaPro$^\dagger$ and DeMo$^\dagger$, our approach demonstrates consistent gains across all metrics. This advantage stems from two key aspects. First, the $\pi$-noise mechanism regularizes unreliable textual semantics into beneficial perturbations rather than treating them as deterministic anchors. Second, the structure-aware prompt modulation enforces geometric consistency across heterogeneous sensors, which is critical in scenes where thermal signatures lack fine-grained textures but retain shape cues. \\
%
\textbf{Multi-Modal Vehicle ReID.}
We further evaluate our method on vehicle ReID benchmarks RGBNT100 and MSVR310, as shown in Table~\ref{tab:multi-spectral vehicle ReID}. The strong performance on the large-scale RGBNT100 dataset validates the generalization capability of our framework. On RGBNT100, Ours$^\dagger$ attains 89.9\% mAP, surpassing TOP-ReID$^*$ by 8.7\% and EDITOR$^*$ by 7.8\%. On the more challenging MSVR310 dataset, our method achieves 65.0\% mAP and 77.2\% Rank-1 accuracy, representing substantial improvements of 18.0\% and 14.8\% over IDEA$^\dagger$. These gains highlight the robustness of our framework in handling real-world complexities. 
Specifically, the CASF component plays a crucial role in MSVR310 by adaptively routing features through experts conditioned on structural context, thereby suppressing background interference from dense urban scenes. 
Meanwhile, the SCM ensures that even when textual prompts contain irrelevant attributes, the injected $\pi$-noise prevents feature corruption and maintains identity fidelity.
These mechanisms allow our model to generalize well across datasets exhibiting diverse degrees of modality heterogeneity and environmental noise.
%
\begin{table}[t]\footnotesize
\caption{Incremental ablation study of the main modules.}
  \centering
    \setlength\tabcolsep{4.5pt}
  \begin{tabular}{cccccccccc}
      \noalign{\hrule height 1pt}
      \multirow{2}{*}{\textbf{Index}} & \multicolumn{3}{c}{\textbf{Modules}} & \multicolumn{2}{c}{\textbf{RGBNT201}} & \multicolumn{2}{c}{\textbf{MSVR310}} &\textbf{FLOPs}  & \textbf{Params} \\
      \cmidrule(r){2-4} \cmidrule(r){5-8}
      & \textbf{SCM} & \textbf{SPA} & \textbf{CASF} 
      &\textbf{mAP} & \textbf{R-1} & \textbf{mAP} & \textbf{R-1} &\textbf{(G)} &\textbf{(M)}\\\hline
  A & \XSolidBrush & \XSolidBrush & \XSolidBrush & 70.3 & 71.9 & 40.4 & 56.0 & 34.27 & 86.41 \\
  B & \XSolidBrush & \XSolidBrush & \CheckmarkBold & 76.5 & 80.3 & 55.8 & 66.3 & 34.29 & 92.73 \\
  C & \XSolidBrush & \CheckmarkBold & \CheckmarkBold & 78.6 & 82.2 & 61.1 & 72.9 & 35.66 & 95.75 \\
  \rowcolor[gray]{0.92}
  D & \CheckmarkBold & \CheckmarkBold & \CheckmarkBold & \textbf{80.6} & \textbf{83.9} & \textbf{65.0} & \textbf{77.2} & 38.68 & 96.23 \\
  \noalign{\hrule height 1pt}
  \end{tabular}
  \label{tab:main_ablation}
\end{table}
%
\begin{table}[t]
  \centering
  \fontsize{9}{9}\selectfont 
  \renewcommand{\arraystretch}{1.4}
  \begin{minipage}[t]{0.49\linewidth}
    \centering
    \caption{Comparison of computational efficiency on RGBNT201.}
    \vspace{1.0mm}%
    \label{tab:flops}
    \setlength\tabcolsep{4.3pt} 
    \begin{tabular}{l|cc}
      \noalign{\hrule height 1pt}
      \textbf{Model} & \textbf{FLOPs (G)} & \textbf{mAP}  \\ \hline
      TOP-ReID$^*$~\cite{wang2024top} & 35.51	 & 72.3  \\
      IDEA$^\dagger$~\cite{wang2025idea} & 43.73 & 80.2  \\
      \rowcolor[gray]{0.92}
      \textbf{Ours$^\dagger$} & \textbf{38.68} & \textbf{80.6}  \\
      \noalign{\hrule height 1pt}
    \end{tabular}
  \end{minipage}
  \hfill 
  \begin{minipage}[t]{0.5\linewidth}
    \centering
    \caption{Comparison of interaction mechanisms in SCM on RGBNT201.}
    \renewcommand{\arraystretch}{1.3}
    \vspace{-1.8mm}%
    \label{tab:scm}
    \setlength\tabcolsep{4.7pt} 
    \begin{tabular}{l|cc}
      \noalign{\hrule height 1pt}
      \textbf{Methods} & \textbf{mAP} & \textbf{R-1}  \\ \hline
      w/o $\pi$-noise modulation & 78.6 & 82.1  \\
      w/ shared noise & 79.2 &80.6 \\
      w/ CA & 79.4	 & 80.3  \\
      \rowcolor[gray]{0.92}
      \textbf{w/ gated CA (Ours)} & \textbf{80.6} & \textbf{83.9}  \\
      \noalign{\hrule height 1pt}
    \end{tabular}
  \end{minipage}
\end{table}
\begin{table}[t]
\centering
\caption{Ablation study on stochastic perturbation mechanisms.}
\label{tab:stochastic}
\setlength{\tabcolsep}{4.7pt}
\begin{tabular}{l|cc|cc|cc|cc|cc}
\noalign{\hrule height 1pt}
\multirow{2}{*}{\textbf{Dataset}} & \multicolumn{2}{c|}{w/o $\pi$} & \multicolumn{2}{c|}{w/ Gauss} & \multicolumn{2}{c|}{w/ uncond.} & \multicolumn{2}{c|}{w/ dropout} & \multicolumn{2}{c}{Ours} \\
\cline{2-11}
& mAP & R-1 & mAP & R-1 & mAP & R-1 & mAP & R-1 & mAP & R-1 \\
\hline
RGBNT201 & 78.6 & 82.1 & 76.1 & 79.6 & 74.8 & 77.5 & 75.7 & 79.2 & \textbf{80.6} & \textbf{83.9} \\
MSVR310 & 61.4 & 73.4 & 60.1 & 73.1 & 59.3 & 70.1 & 60.9 & 73.3 & \textbf{65.0} & \textbf{77.2} \\
\noalign{\hrule height 1pt}
\end{tabular}
\end{table}
\\
\textbf{Effect of Key Modules.}
As shown in Table~\ref{tab:main_ablation}, we conduct ablation studies on RGBNT201 and MSVR310 to evaluate the contribution of our three core components proposed in Sec.~\ref{sec:method}, all of which play a critical role in boosting overall performance.
Taking RGBNT201 as an example, Exp. B validates the role of CASF in mitigating local noise. By routing features through sparse experts conditioned on structural context rather than fusing raw patch tokens directly, CASF avoids propagating unreliable details from cluttered backgrounds. This design yields a substantial improvement of +6.2\% mAP and +8.4\% Rank-1, confirming that adaptive fusion guided by clean structural signals is critical for robust identity representation.
Building on this, Exp. C demonstrates that the SPA further refines cross-modal alignment. SPA injects DINOv3-derived geometric priors via learnable prompts and aligns them across modalities through a shared adapter. The resulting gains (78.6\% mAP, 82.2\% Rank-1) highlight that explicit structural guidance compensates for texture degradation, enabling more consistent feature matching under modality heterogeneity.
Finally, Exp. D shows that the SCM unlocks the full potential of textual semantics. Instead of treating text as a fixed anchor, SCM leverages  Positive-Incentive Noise ($\pi$-noise) to regularize textual uncertainty into a beneficial perturbation. This leads to the best overall performance (80.6\% mAP, 83.9\% Rank-1), proving that semantic signals, when properly modulated, can actively enhance discriminability rather than introduce bias. Similar consistent improvements are also observed on MSVR310.\\
%
%
\textbf{Computational Efficiency and FLOPs Analysis.}
We analyze the computational efficiency of our method on RGBNT201 in Table~\ref{tab:main_ablation} and Table~\ref{tab:flops}.  
Our model achieves the best performance (80.6\% mAP) with 38.68\, GFLOPs.
Specifically, our text branch configuration and FLOPs calculation follow the protocol established in prior work~\cite{wang2025idea,xu2026stmi}. The reduced FLOPs compared to IDEA arise from our shared text encoder across all three modalities, whereas IDEA employs a separate text encoder for each spectral modality.  
This efficiency stems from our lightweight cross-modal interaction design: the majority of computation is dominated by the shared CLIP backbone, and we deliberately avoid introducing additional heavy fusion modules or modality-specific transformers. 
\begin{table}[t]
  \centering
  \fontsize{8}{8}\selectfont 
  \renewcommand{\arraystretch}{1.4}
  \begin{minipage}[t]{0.55\linewidth}
    \centering
    \caption{Comparison of prompt configurations in SPA on RGBNT201.}
    \label{tab:spa}
    \vspace{-2.3mm}
    \setlength\tabcolsep{4pt} 
    \begin{tabular}{l|cc}
      \noalign{\hrule height 1pt}
      \textbf{Methods} & \textbf{mAP} & \textbf{R-1}  \\ \hline
      w/ linear & 78.0 &80.5  \\
      w/o adapter & 79.1 & 81.3  \\
      w/o prompt  & 77.0	 & 81.2  \\
      prompt w/o DINO init & 78.0	 & 81.3  \\
      \rowcolor[gray]{0.92}
      \textbf{prompt w/ DINO init (Ours)} & \textbf{80.6} & \textbf{83.9}  \\
      \noalign{\hrule height 1pt}
    \end{tabular}
  \end{minipage}
  \hfill 
  \begin{minipage}[t]{0.43\linewidth}
    \centering
    \caption{Comparison of routing strategies in CASF on RGBNT201.}
    \vspace{1.6mm}
    \label{tab:casf}
    \setlength\tabcolsep{4.5pt} 
    \begin{tabular}{l|cc}
      \noalign{\hrule height 1pt}
      \textbf{Methods} & \textbf{mAP} & \textbf{R-1}  \\ \hline
      w/ CA & 78.1 & 79.2  \\
      w/o MoE & 78.5 & 80.6  \\
      input w/ $[f_R,f_N,f_T]$  & 78.6	 & 81.8  \\
      \rowcolor[gray]{0.92}
      \textbf{input w/ $\tilde{f}_g$ (Ours)} & \textbf{80.6} & \textbf{83.9}  \\
      \noalign{\hrule height 1pt}
    \end{tabular}
  \end{minipage}
\end{table}
\subsection{Further Analysis}
%
\textbf{Semantic Cross-Modal Modulator.}
We present a detailed ablation study on the internal design choices of SCM in Table~\ref{tab:scm}.
The results show that removing $\pi$-noise modulation (w/o $\pi$-noise modulation) leads to 78.6\% mAP and 82.1\% Rank-1, which are 2.0\% and 1.8\% lower than our full model. 
We also test a variant with a shared $\pi$-noise across modalities (w/ shared noise). While it benefits from noise-induced perturbation, it cannot adapt modulation to each modality’s characteristics.
Moreover, using standard cross-attention (w/ CA) instead of the gated mechanism yields inferior performance, indicating that gating conditioned on visual global features enables more effective text-image feature fusion. \\
\textbf{Stochasticity Analysis.} 
Table~\ref{tab:stochastic} verifies that the gain of $\pi$-noise stems from task-aware structure rather than mere stochasticity. 
Compared to the deterministic baseline (w/o $\pi$), naive Gaussian noise (w/ Gauss), unconditional sampling (w/ uncond.), and dropout (w/ dropout) all degrade performance, confirming that unstructured randomness harms the aligned CLIP feature space. 
In contrast, our semantically-conditioned $\pi$-noise achieves the best results.\\
%
\textbf{Structure-Aware Prompt Adapter.}
To examine the effectiveness of structural prior injection in SPA, we compare different designs in Table~\ref{tab:spa}. 
Replacing the adapter with a simple linear layer (w/ linear) leads to worse performance.
This indicates that a naive linear transformation fails to properly modulate features with the structural prior and may even introduce harmful interference.
Bypassing the adapter and directly using the prior for routing (w/o adapter) reduces mAP and Rank-1 by 1.5\% and 2.6\%, respectively, highlighting the adapter's role in stable feature transformation. Using only CLIP patch tokens without structural prompts (w/o prompt) results in 77.0\% mAP, while initializing prompts randomly instead of with DINOv3 (prompt w/o DINO init) achieves 78.0\% mAP, both below the full model (Ours). This confirms that lightweight DINOv3 initialization effectively injects structural information.
%
\begin{table}[t]
\footnotesize
\centering
\fontsize{8}{8}\selectfont 
\renewcommand{\arraystretch}{1.3}
\caption{Arbitrary modality-missing patterns on RGBNT201 and MSVR310. R, N, and T respectively denote RGB, NIR, and TIR. M(X) means missing modality X.}
\label{tab:modality_missing}
\setlength{\tabcolsep}{0.5pt}
\begin{tabular}{c|c|cccccccccccc|cc}
\noalign{\hrule height 1pt}
\multirow{2}{*}{\textbf{Dataset}} & \multirow{2}{*}{Method} & \multicolumn{2}{c}{M(R)} & \multicolumn{2}{c}{M(N)} & \multicolumn{2}{c}{M(T)} & \multicolumn{2}{c}{M(RN)} & \multicolumn{2}{c}{M(RT)} & \multicolumn{2}{c|}{M(NT)} & \multicolumn{2}{c}{Avg.} \\
\cline{3-16}
& & mAP & R-1 & mAP & R-1 & mAP & R-1 & mAP & R-1 & mAP & R-1 & mAP & R-1 & mAP & R-1 \\
\hline
\multirow{2}{*}{{{RGBNT201}}} & DeMo & 63.3 & 65.3 & \textbf{72.6} & \textbf{75.7} & 56.2 & 54.1 & \textbf{45.6} & \textbf{46.5} & 26.3 & 24.9 & \textbf{40.3} & 38.5 & 50.7 & 50.8 \\
& Ours & \textbf{64.5} & \textbf{65.9} & 71.5 & 74.4 & \textbf{62.9} & \textbf{62.7} & 43.2 & 46.2 & \textbf{37.8} & \textbf{37.7} & 39.5 & \textbf{41.2} & \textbf{53.2} & \textbf{54.7} \\
\hline
\multirow{2}{*}{{{MSVR310}}} & DeMo & 35.9 & 50.3 & \textbf{42.0} & \textbf{58.4} & 43.1 & \textbf{59.7} & \textbf{15.3} & \textbf{28.8} & 27.2 & 43.5 & 35.3 & 49.6 & 33.1 & 48.4 \\
& Ours & \textbf{43.8} & \textbf{57.2} & 40.1 & 54.7 & \textbf{45.0} & 58.7 & 11.3 & 21.3 & \textbf{32.2} & \textbf{49.1} & \textbf{41.6} & \textbf{56.9} & \textbf{35.7} & \textbf{49.6} \\
\noalign{\hrule height 1pt}
\end{tabular}
\end{table}
%
%
\begin{figure}[t]
\centering
\includegraphics[width=1.0\linewidth]{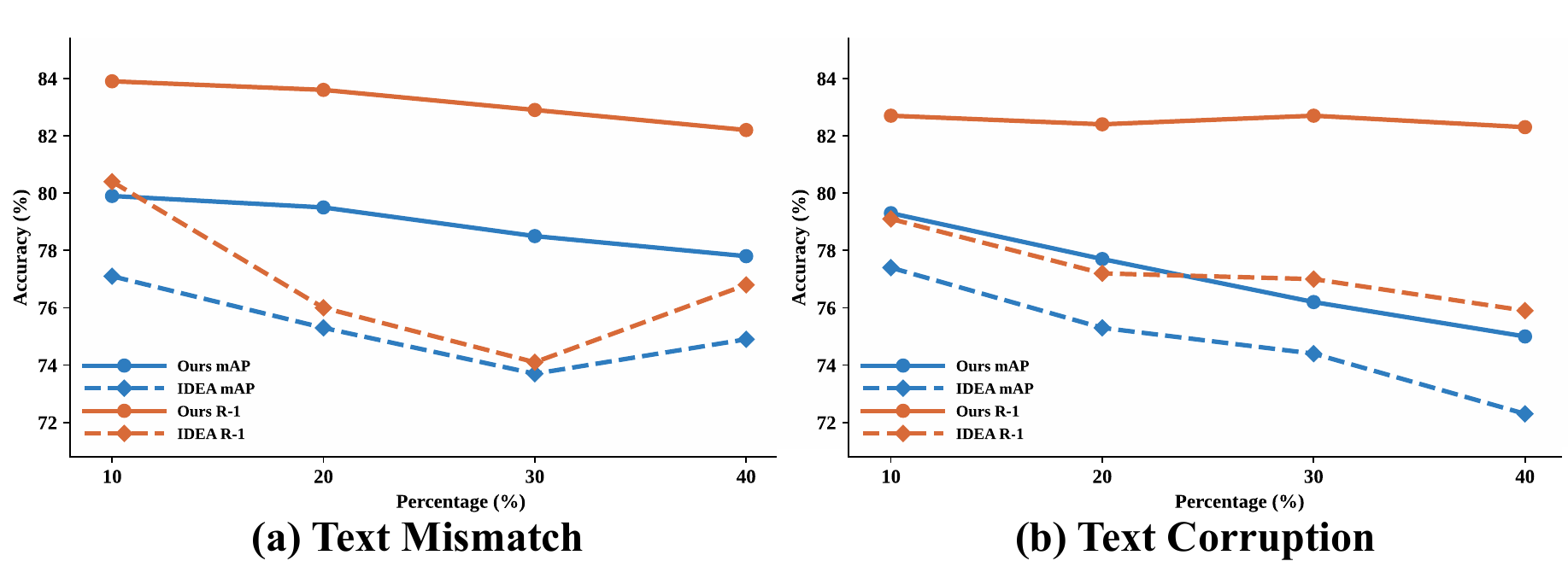}
\caption{Comparison under text corruption and mismatch on RGBNT201.}
\label{fig:corruption}
\end{figure}
%
\begin{table}[t]
\centering
\fontsize{7}{7}\selectfont 
    \begin{minipage}[t]{0.31\textwidth}
        \centering
        \setlength\tabcolsep{2.5pt}
        \renewcommand{\arraystretch}{1.4}
        \caption{Number of GeM pooling branches on RGBNT201.}
        \begin{tabular}{c|cccc}
            \noalign{\hrule height 1pt}
            \textbf{$|N'_t|$} & \textbf{mAP} & \textbf{R-1}  & \textbf{R-5} & \textbf{R-10}\\ \hline
            1 & 79.7 & 83.9  & 92.0	 & 93.8\\
            \rowcolor[gray]{0.92}
            {2} & {80.6} & {83.9}  & 91.6	 & 93.4\\
            3  & 80.4	 & 83.3  & 91.0	 & 93.4\\
            4  & 80.3	 & 83.9  & 91.3	 & 94.6\\
            5  & 79.1	 & 81.3  & 89.2	 & 92.0 \\
            \noalign{\hrule height 1pt}
        \end{tabular}
        \label{tab:gem}
    \end{minipage}
    \hfill
    \begin{minipage}[t]{0.68\textwidth}
        \centering
        \setlength\tabcolsep{4pt}
        \renewcommand{\arraystretch}{1.3}
        \caption{Comparison of the number of trainable parameters among different methods.}
        \vspace{-1.2mm}
        \begin{tabular}{r|c|cccc}
            \noalign{\hrule height 1pt}
            \multicolumn{1}{c|}{\multirow{2}{*}{\textbf{Methods}}} &
            \multicolumn{1}{c|}{\multirow{2}{*}{\textbf{Params (M)}}} & \multicolumn{2}{c}{\textbf{RGBNT201}} & \multicolumn{2}{c}{\textbf{MSVR310}} \\\cline{3-6}
            && \textbf{mAP} & \textbf{R-1} & \textbf{mAP} & \textbf{R-1} \\
            \hline
            CLIP Baseline$^\dagger$ &86.41& 70.3 &71.9 & 40.4 & 56.0 \\
            TOP-ReID$^*$~\cite{wang2024top} &324.53 &72.3 & 76.6 & 35.9 & 44.6 \\
            EDITOR$^*$~\cite{zhang2024magic} &118.55& 66.5 & 68.3 &39.0 & 49.3\\
            WTSF-ReID$^*$~\cite{yu2025wtsf} &143.60& 67.9 &72.2 & 39.2 & 49.1 \\
            DeMo$^\dagger$~\cite{wang2025decoupled} &98.79& 79.0 &82.3 & 49.2 & 59.8 \\
            \rowcolor[gray]{0.92}
        ${\textbf{Ours}}^\dagger$&\textbf{96.23}&\textbf{80.6} 	&\textbf{83.9} &\textbf{65.0}	&\textbf{77.2} \\
            \noalign{\hrule height 1pt}
        \end{tabular}
        \label{tab:params}
    \end{minipage}
\end{table}
\\
\textbf{Context-Aware Sparse Fusion.}
As shown in Table~\ref{tab:casf}, we evaluate the routing-based fusion mechanism in CASF by comparing it against two variants: (1) replacing the MoE router with a linear fusion layer ({w/o MoE}), and (2) using cross-attention to directly fuse modality-specific features ({w/ CA}). 
Both variants underperform the full model, indicating that direct feature blending without selective routing introduces cross-modal interference and degrades discriminative cues. 
Additionally, restricting the MoE inputs to modality-specific global tokens ({w/} $[f_R,f_N,f_T]$) while excluding incentive noise from routing yields lower mAP and Rank-1. 
This validates our routing design and the efficacy of incorporating noise sampled from semantic-aware distributions.\\
\textbf{Robustness to Missing Modalities.} 
Table~\ref{tab:modality_missing} evaluates arbitrary modality-missing patterns on RGBNT201 and MSVR310. 
Our method achieves a higher average performance than DeMo. 
Notably, under the challenging M(RT) setting, our method achieves 37.8\% mAP while DeMo achieves 26.3\% on RGBNT201. 
This demonstrates stronger compensation when critical modalities are absent.\\
%
\textbf{Robustness to Textual Degradation.}
We further evaluate robustness under two textual degradation protocols: mismatch (random description swaps) and corruption (partial attribute replacement with Unknown). 
Figure~\ref{fig:corruption} shows that our method consistently outperforms IDEA across all ratios. 
Under 20\% corruption, the baseline drops 18.6\% mAP while ours drops only 3.6\%, confirming that $\pi$-noise disproportionately benefits high-noise scenarios.\\
\textbf{Number of GeM Pooling Branches.}
Table~\ref{tab:gem} reports the effect of varying the number of pooled tokens $|N'_t|$ in the text feature $f_t$. The model achieves the best mAP and Rank-1 at $|N'_t| = 2$, with marginal drops for other values. This indicates that two tokens strike a good balance between semantic preservation and compactness, and the model is robust to the pooling granularity.\\
\textbf{Trainable Parameter Count.}
Table~\ref{tab:main_ablation} shows the parameter increase from each module, and Table~\ref{tab:params} lists the total.
Compared to the CLIP baseline$^\dagger$, our model adds only a few parameters while gaining +10.3\% mAP on RGBNT201 and +24.6\% mAP on MSVR310.
Our approach also surpasses established methods such as EDITOR$^*$~\cite{zhang2024magic} and DeMo$^\dagger$~\cite{wang2025decoupled} while using fewer trainable parameters.
\begin{figure}[t]
  \centering
  \includegraphics[width=\linewidth]{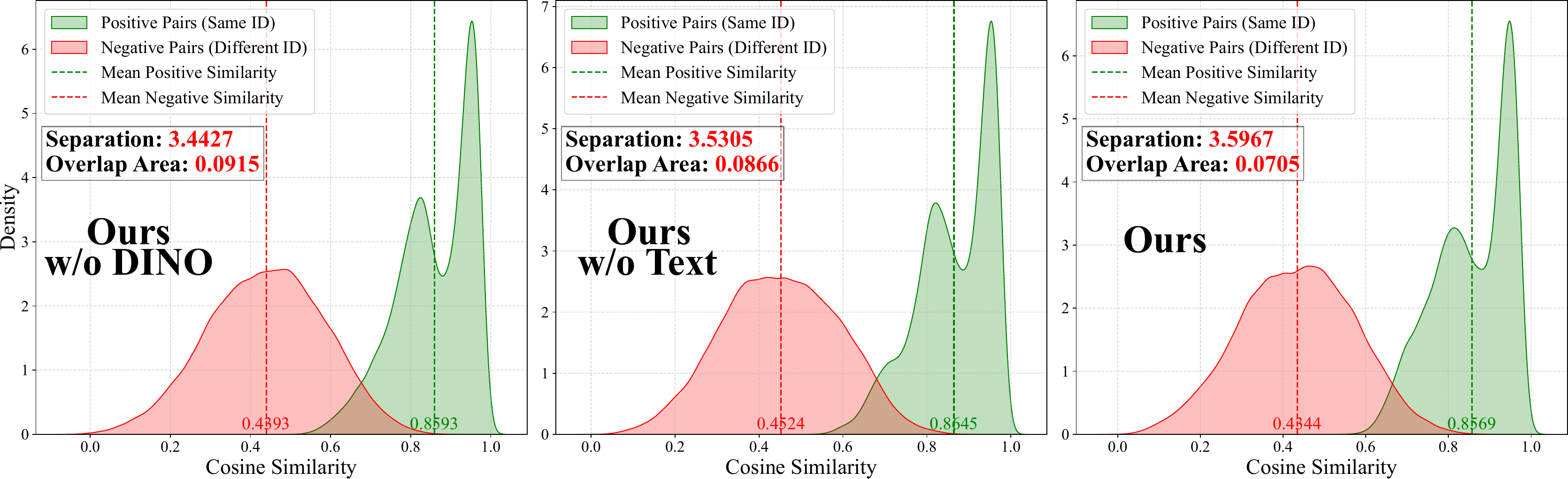}
    \caption{Cosine similarity distributions of positive and negative feature pairs on RGBNT201 for our full model and ablated variants (w/o Text and w/o DINO).}
  \label{fig:cosine}
\end{figure}
\begin{figure}[t]
  \centering
  \includegraphics[width=\linewidth]{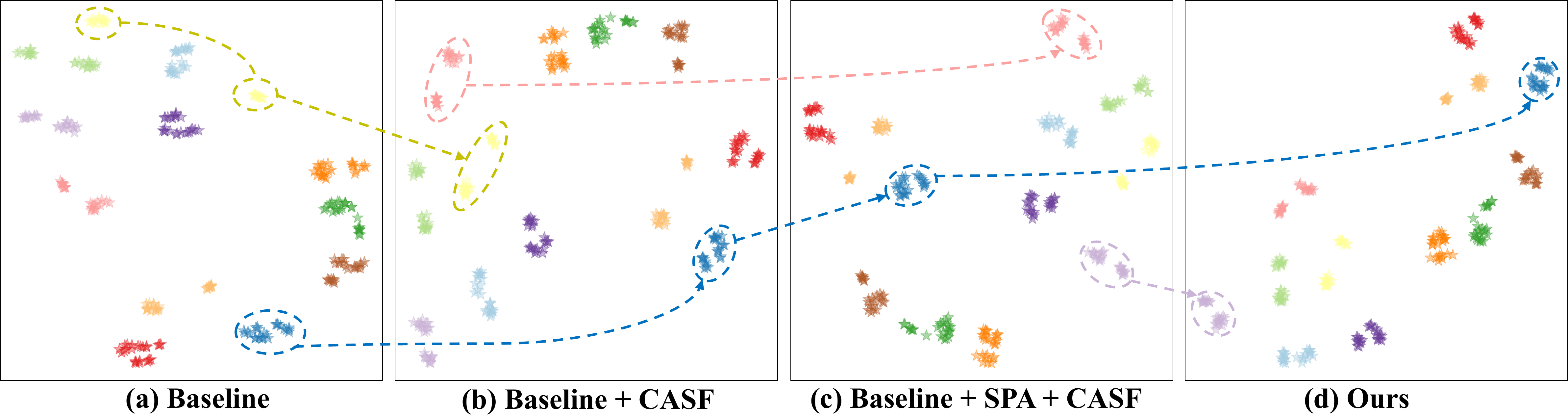}
   \caption{Visualization of multi-modal ReID features on RGBNT201 using t-SNE~\cite{van2008visualizing}.
   Ours means the full model (baseline+CASF+SPA+SCM).
   }
  \label{fig:tsne}
\end{figure}
\begin{figure}[t]
  \centering
  \includegraphics[width=\linewidth]{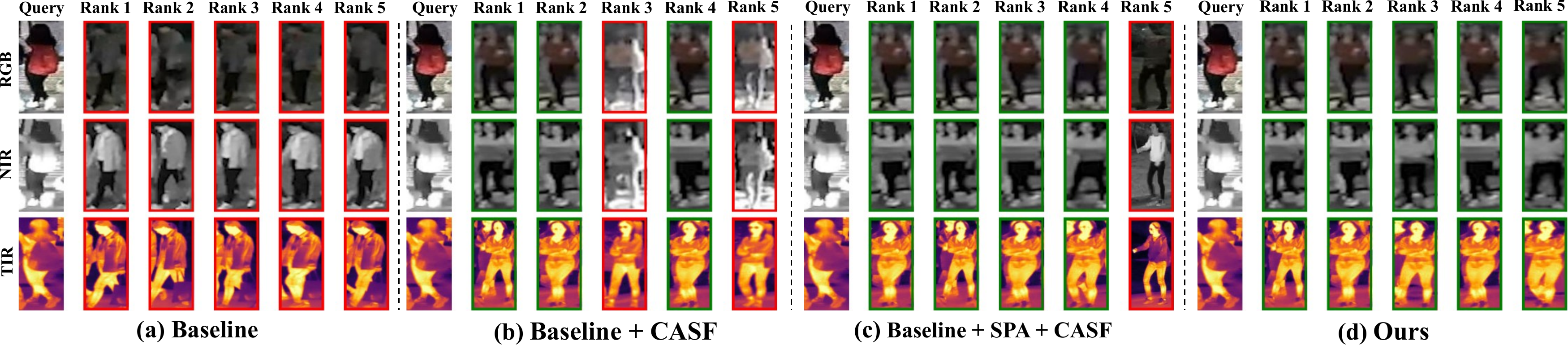}
   \caption{Rank list comparison between the baseline and our approach.}
  \label{fig:rank_list}
\end{figure}
\subsection{Visualization Analysis}
%
\textbf{Cosine Similarity Distributions.}
To examine the impact of text-guided noise and structural priors on feature similarity distributions, we present the cosine similarity histograms in Figure~\ref{fig:cosine}.
We adopt two lightweight metrics to quantify feature discriminability: the separation metric (higher is better) and the overlap area between positive and negative similarity distributions (lower is better). A higher separation metric (\textbf{Separation} $\uparrow$) and a lower overlap area (\textbf{Overlap Area} $\downarrow$) jointly indicate stronger intra-class compactness and inter-class separability. 
%
Ablations without learnable structural priors (w/o DINO) or without textual guidance (w/o Text) both lead to reduced separation and increased overlap, confirming the effectiveness of our design. 
Specifically, the results show that incorporating DINOv3-initialized structural priors and text-guided beneficial noise improves feature separation, yielding a separation metric of 3.5967 and an overlap area of 0.0705.
When combined, these components produce the clearest inter-class separation and intra-class compactness among all variants, confirming their complementary roles in shaping a discriminative feature space.
\\
\textbf{Multi-Modal Feature Distributions.}
Figure~\ref{fig:tsne} displays t-SNE~\cite{van2008visualizing} embeddings of features from several model configurations. Comparing Figure~\ref{fig:tsne}(a) and (b), the addition of CASF reduces ambiguity in regions such as the yellow circle, where samples from the same identity exhibit less overlap with neighboring classes. Further comparisons between (b) and (c), and between (c) and (d), reveal that clusters in the pink and purple circles become increasingly distinct after introducing SPA and SCM, respectively. Additionally, the cluster highlighted in blue becomes progressively more compact as each component is added. These observations reflect a consistent refinement in feature geometry across stages.
\\
\textbf{Rank List Comparison.}
Figure~\ref{fig:rank_list} compares cross-camera retrieval results from the baseline and our method. Our approach returns more accurate rank lists, with relevant matches consistently ranked higher and fewer irrelevant items in the top positions. In contrast, the baseline shows greater variability and includes more false matches, especially under challenging conditions such as pose changes or low-quality queries. This qualitative comparison supports the conclusion that our model learns more discriminative and robust cross-modal representations. 
%
\section{Conclusion}
In this paper, we present a novel multi-modal object ReID framework that rethinks textual noise and structural context as constructive signals for cross-modal learning.  
The Semantic Cross-Modal Modulator (SCM) leverages $\pi$-noise for semantics-driven interaction between vision and language.  
The Structure-Aware Prompt Adapter (SPA) injects learnable geometric priors via prompts to align cross-modal structures.  
The Context-Aware Sparse Fusion (CASF) leverages structural context to guide sparse fusion while preserving identity discrimination from noisy local features.  
Experiments on standard benchmarks validate the effectiveness and generalization capability of our approach.\\
\textbf{Acknowledgements.}  
This work was supported in part by the National Natural Science Foundation of China under Grant 62476041.




%
%
\bibliographystyle{splncs04}
\bibliography{main}
\end{document}